\pdfoutput=1

\documentclass[11pt]{article}

\usepackage[preprint]{acl}

\usepackage{times}
\usepackage{latexsym}
\usepackage{booktabs}
\usepackage{multirow}
\usepackage{graphicx}
\usepackage[table]{xcolor}
\usepackage{colortbl}
\usepackage{amsmath}
\usepackage{amsfonts}
\definecolor{datablue}{HTML}{6699FF}
\usepackage{arydshln} 
\usepackage{amsthm}
\usepackage{thmtools}
\usepackage{hyperref}
\usepackage{mathtools}
\newcommand{\grayrow}[1]{\textcolor[gray]{0.55}{#1}}
\newcommand{\bluerow}[1]{\textcolor{blue!55}{#1}}
\usepackage{amsmath, amssymb, amsthm}
\newtheorem{proposition}{Proposition}
\definecolor{tabblue}{HTML}{1f77b4}
\definecolor{tabred}{HTML}{d62728}
\usepackage[T1]{fontenc}

\usepackage[utf8]{inputenc}

\usepackage{microtype}

\usepackage{inconsolata}

\usepackage{graphicx}

\title{ELLA: Efficient Lifelong Learning for Adapters in Large Language Models}

\vspace{-50pt}
\author{Shristi Das Biswas\thanks{Work conducted at Amazon.}\\
  Purdue University \\
  \\\And
  Yue Zhang \\
  AWS \\
  \\\And
  Anwesan Pal \\
  AWS AI Labs \\
  \\\And
  Radhika Bhargava \\
  AWS \\
  \\\And
  Kaushik Roy \\
  Purdue University \vspace{-20pt}
  }
\vspace{-50pt}

\begin{document}
\vspace{-27pt}
\maketitle
\vspace{-27pt}
\renewcommand{\thefootnote}{\fnsymbol{footnote}}
\footnotetext[1]{\url{https://sites.google.com/view/ella-llm/home}.}
\renewcommand{\thefootnote}{\arabic{footnote}}
\begin{abstract}
\vspace{-5pt}
Large Language Models (LLMs) suffer severe catastrophic forgetting when adapted sequentially to new tasks in a continual learning (CL) setting. Existing approaches are fundamentally limited: replay-based methods are impractical and privacy-violating, while strict orthogonality-based methods collapse under scale: each new task is projected onto an orthogonal complement, progressively reducing the residual degrees of freedom and eliminating forward transfer by forbidding overlap in shared representations. In this work, we introduce ELLA, a training framework built on the principle of selective subspace de-correlation. Rather than forbidding all overlap, ELLA explicitly characterizes the structure of past updates and penalizes alignments along their high-energy, task-specific directions, while preserving freedom in the low-energy residual subspaces to enable transfer. Formally, this is realized via a lightweight regularizer on a single aggregated update matrix. We prove this mechanism corresponds to an anisotropic shrinkage operator that bounds interference, yielding a penalty that is both memory- and compute-constant regardless of task sequence length. ELLA requires no data replay, no architectural expansion, and negligible storage. Empirically, it achieves state-of-the-art CL performance on three popular benchmarks, with relative accuracy gains of up to $9.6\%$ and a $35\times$ smaller memory footprint. Further, ELLA scales robustly across architectures and 
actively enhances the model's zero-shot generalization performance on unseen tasks, establishing a principled and scalable solution for constructive lifelong LLM adaptation.$^\star$
\end{abstract}

\section{Introduction}
Large Language Models (LLMs) have demonstrated remarkable generalization capabilities across a wide range of downstream tasks, largely attributed to their large-scale pretraining on diverse corpora~\cite{brown2020language, touvron2023llama,achiam2023gpt}. As LLMs are increasingly deployed in real-world applications, they must be continuously adapted to evolving user needs and task distributions for long-term practical deployment. This setting, commonly studied under the paradigm of continual learning (CL), requires models to acquire new knowledge sequentially without costly full-retraining~\cite{ruvolo2013ella}. However, sequential finetuning of LLMs remains highly susceptible to catastrophic forgetting (CF) - the tendency to overwrite prior knowledge when new tasks are introduced~\cite{mccloskey1989catastrophic}, and loss of plasticity - deterioration in the ability to learn new information over time~\cite{dohare2021continual}. These challenges are especially pronounced in rehearsal-free settings, where previously seen data cannot be stored due to privacy or storage constraints~\cite{chaudhry2019continual}.

A promising line of work leverages parameter-efficient fine-tuning (PEFT) strategies, such as Low-Rank Adaptation (LoRA)~\cite{hu2022lora}, to reduce the overhead of task-specific adaptation. Adapter-based approaches restrict updates to a small set of trainable parameters while leaving the base model frozen, making them a natural fit for lower-compute CL~\cite{qin2021lfpt5, song2023conpet}. Yet, even with such modularity, sequential adapter training often yields severe forgetting when past knowledge is not revisited~\cite{wang2024inscl}. Prior efforts to address this include expanding model capacity~\cite{wang2024rehearsal}, isolating weights to reduce interference~\cite{aljundi2017expert, li2019learn, wang2023rehearsal}, subspace orthogonality~\cite{wang2023orthogonal, liao2025data}, or gradient projection from previous tasks~\cite{qiao2024learn}. While effective to varying degrees, these often either limit forward knowledge transfer from past tasks, add storage overhead, or overlook activation-level interference, where forgetting actually manifests~\cite{ke2021achieving}.

A truly effective continual learning framework should balance knowledge retention with representation reuse, preventing harmful interference while allowing useful subspaces to be shared for new tasks~\cite{wu2024continual}. In practice, not all overlap is harmful -- low-magnitude directions from past tasks may encode generic linguistic or semantic patterns that can accelerate learning on future tasks if safely reused. Unfortunately, existing CL methods either eliminate all overlap or fuse past knowledge through heavyweight mechanisms like controller networks or rank-conditioned fusion, which limits scalability and increases complexity~\cite{zhao2024sapt, liu2023vida, liao2025data}.

In this work, we propose ELLA, a simple and scalable CL framework for LLMs that mitigates forgetting without relying on replay buffers, parameter storage overheads or additional routing heuristics. ELLA introduces a subspace-aware regularization strategy that operates directly in weight space: we track the representational subspaces induced by past adapters and penalize updates that cause the new task's adapter to align too closely with them. 
This cross-task de-correlation simultaneously encourages task-specific specialization by preserving prior representational geometry to reduce interference, while permitting the reuse of low-magnitude directions to enable forward transfer.
As we formally prove in Appendix~\ref{app:proof}, this selective regularization corresponds to an anisotropic shrinkage operator that provably bounds interference, providing a theoretical foundation for ELLA's ability to balance stability and plasticity.

Through extensive experiments on the Standard CL Benchmark~\cite{zhang2015character}, Long Sequence Benchmark~\cite{razdaibiedina2023progressive} and TRACE~\cite{wang2023trace}, we demonstrate that ELLA achieves state-of-the-art performance while scaling effectively across model sizes (from $770M$ to $8B$). Furthermore, ELLA generalizes well across architecture families (e.g., T5~\cite{raffel2020exploring}, LLaMA~\cite{touvron2023llama}), and unlike prior work, uniquely improves generalization to unseen tasks. ELLA does not rely on task identities during inference, and is hence naturally compatible with the instruction-tuning paradigm~\cite{wang2022task}, thereby preserving the generalization capabilities of LLMs in zero-shot and open-ended settings. Notably, our method can also be seamlessly integrated with existing continual learning methods to further enhance their effectiveness, without requiring additional supervision or auxiliary components. Moreover, ELLA is architecture-agnostic, 
requires no access to past data, and introduces negligible computational or memory overhead. In summary, our contributions are as follows:
\begin{itemize}
\item 
We propose ELLA, a replay-free plug-and-play CL framework for LLMs that balances plasticity and mitigates CF via subspace-aware regularization, and we provide a formal theoretical analysis of its properties.
\item We provide extensive empirical evidence that ELLA establishes a new state-of-the-art in both performance and efficiency across three popular CL benchmarks, decisively outperforming prior methods without task labels, replay, or additional overhead.
\item We show that ELLA scales effectively across architectures and maintains strong generalization to unseen tasks, 
highlighting its practical advantages for real-world deployment.
\end{itemize}

\section{Related Works}
\textbf{Continual Learning} \hspace{5pt}
The primary challenge in CL is to balance model stability (resisting catastrophic forgetting) with plasticity (acquiring new knowledge). Prior approaches to address this problem can be broadly categorized by their core mechanism, each presenting a distinct set of trade-offs.
\textbf{(i) Rehearsal-based methods} store a subset of past data and interleave it during subsequent training phases. These include strategies such as experience replay~\cite{riemer2018learning} or constrained optimization~\cite{aljundi2017expert, chaudhry2019continual, he2024seekr} that jointly train on current and previous samples. While empirically strong, this approach is often impractical due to significant storage overhead and potential violations of data privacy constraints.
\textbf{(ii) Regularization-based methods} add penalty terms to the loss that restrict updates to parameters critical for old tasks~\cite{du2024unlocking, li2017learning, de2019episodic}. For instance,~\cite{kirkpatrick2017overcoming} slows updates on important weights, while~\cite{farajtabar2020orthogonal} constrains new updates to be orthogonal to gradients of old tasks. These methods depend on importance metrics that are often brittle and prevent any forward transfer, limiting adaptability to new tasks.
\textbf{(iii) Architecture-based methods} reduce interference through task-specific modules or by expanding model size~\cite{li2019learn, wang2023rehearsal}. While~\cite{razdaibiedina2023progressive} appends a new learned soft prompt per task, ~\cite{qin2021lfpt5} utilizes a large
soft prompt that is continuously trained on all tasks. While this effectively isolates task knowledge, it typically results in a linear growth in parameter count with the number of tasks, posing significant scalability challenges and often requiring explicit task labels at inference.

\textbf{Parameter-Efficient Continual Learning} \hspace{5pt}
Parameter-Efficient Fine-Tuning (PEFT) has emerged as a highly promising substrate for CL, with Low-Rank Adaptation (LoRA)~\cite{hu2022lora} being a particularly effective technique. However, naïvely applying LoRA sequentially leads to significant forgetting owing to \textit{interference across tasks}. As tasks arrive sequentially, the adapter weights $A_t$, $B_t$ learned for a new task $t$ are trained from scratch without awareness of previous tasks' LoRA subspaces. Consequently, overlapping update directions in parameter space may inadvertently override past knowledge, especially under limited parameter budgets. Without coordination across tasks, these LoRA components can destructively interfere in spaces crucial for previous tasks, degrading performance on earlier tasks while optimizing for the current one. Recent works in CL literature addressing this issue primarily falls into two schools of thought. 

The first approach applies strict orthogonality to enforce zero overlap between the LoRA subspaces of different tasks~\cite{wang2023orthogonal, qiao2024learn}. While this hard constraint effectively mitigates interference, it imposes an overly restrictive inductive bias. By precluding any form of representation sharing, it severely inhibits beneficial forward transfer and prevents reuse of low-importance shared space across related tasks. Worse still, as tasks accumulate, the available orthogonal space quickly diminishes, leading to progressively less capacity for future learning. This rigid decoupling causes inefficient use of adapter capacity, hindering the goal of cumulative learning. Furthermore, storing all past LoRA weights incurs high memory overhead that scales linearly with the number of tasks, further limiting scalability.

The second approach involves more complex fusion and replay mechanisms. Methods like DATA~\cite{liao2025data} or Recurrent KIF~\cite{feng2025recurrent} often rely on replay buffers, multiple LoRA adapters per task, or intricate gating mechanisms to learn how to compose adapters. These methods reintroduce significant architectural complexity and computational overhead, sacrificing the very simplicity and efficiency that make PEFT an attractive paradigm for CL in the first place.

Differently, ELLA charts a third path. The prevailing paradigms are caught in a dilemma: rehearsal is impractical, naïve regularization is ineffective, architectural expansion is inefficient, orthogonality is too restrictive, and complex fusion is costly. In contrast, ELLA abandons the assumption that all interference must be uniformly suppressed and introduces the more effective principle of managing interference selectively. By regularizing only the high-magnitude, task-specific directions of the adapter space while permitting the reuse of low-magnitude, generalizable features, ELLA achieves a superior balance of stability and plasticity without the trade-offs that limit prior work, and integrates seamlessly into LoRA-based workflows, enabling scalable and robust CL adaptation for LLMs.

\section{Method}
\subsection{Problem Formulation}
\textbf{Setup.} 
In the supervised CL setting, a model encounters a stream of tasks $\{1, \dots, \mathcal{T}\}$ sequentially, where each task $t = \{(x^t_i, y^t_i)\}_{i=1}^{n_t}$ contains a labeled dataset $x^t_i \in \mathcal{X}_t$ and $y^t_i \in \mathcal{Y}_t$. Given a prediction model $h_\Theta$ with parameters $\Theta$, the objective is to maximize total log-likelihood over all tasks:
\vspace{-9pt}
\begin{equation}
\max_{\Theta} \sum_{k=1}^{\mathcal{T}} \sum_{(x, y) \in \mathcal{X}_k, \mathcal{Y}_k} \log p_\Theta(y \mid x)
\vspace{-9pt}
\end{equation}
Here, we consider a more challenging setting for ELLA: \textit{rehearsal-free CL with task-agnostic inference}. (i) 
The model cannot store or access data from past tasks when learning a new one. (ii) At test time, the model must make predictions without knowing which task an input belongs to.
\begin{figure*}[!t]
\centering
\includegraphics[width=0.98\textwidth]{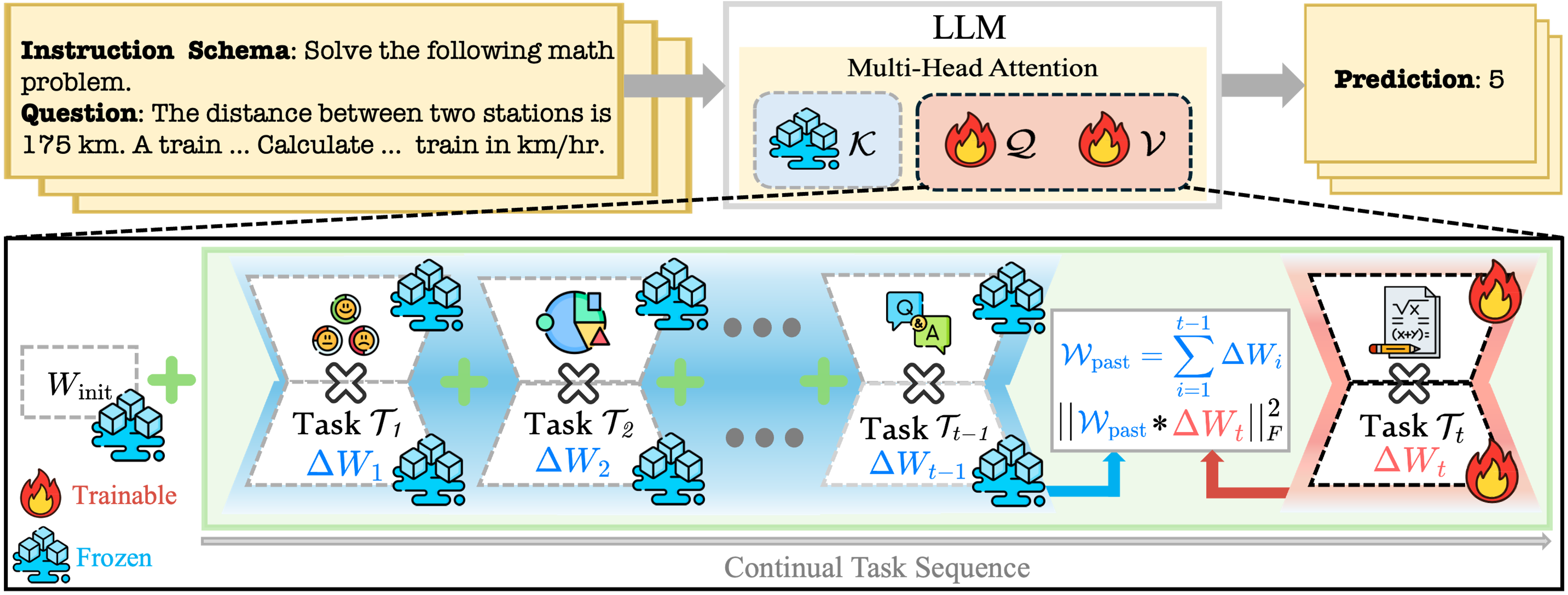} 
\vspace{-5pt}
\caption{ELLA mitigates interference in continual LoRA training by accumulating past low-rank updates $\mathcal{W}_{\text{past}}$ and applying an energy-based alignment penalty $|| \Delta W_t * \mathcal{W}_{\text{past}} ||_F^2$ to discourage overlap in high-magnitude, task-specific directions. This selective regularization enables parameter reuse in less-used subspaces, achieving a better trade-off between plasticity and stability without requiring task labels, data replay, or architectural modifications.}
\vspace{-6pt}
\label{fig:method}
\end{figure*}

\textbf{Low-Rank Adaptation (LoRA).} Our method builds on LoRA~\cite{hu2022lora}, a PEFT technique that leverages the observation that fine-tuning large pre-trained models (PTMs) can be effectively performed within a low-dimensional subspace~\cite{aghajanyan2020intrinsic}. Given a frozen pre-trained weight matrix $W_{\mathrm{init}} \in \mathbb{R}^{d \times k}$, LoRA introduces a low-rank update $\Delta W = AB$, where $A \in \mathbb{R}^{d \times r}$ and $B \in \mathbb{R}^{r \times k}$ with $r \ll \min(d, k)$. The original weights $W_{\mathrm{init}}$ remain unchanged during training, and only the low-rank matrices $A$ and $B$ are optimized. During the forward pass, the original operation $h = W_{\mathrm{init}} x$ becomes:
\vspace{-4pt}
\begin{equation}
    h = W_{\mathrm{init}} x + \Delta W x = W_{\mathrm{init}} x + ABx,
\vspace{-4pt}
\end{equation}
This design reduces memory and compute overhead significantly, while achieving competitive performance to full fine-tuning across model families and benchmarks~\cite{houlsby2019parameter}.

\subsection{Subspace-Aware Continual Adaptation}
\label{sec:subsapace}
We propose ELLA, a simple yet effective extension to LoRA, as a CL framework that balances \textit{plasticity} and \textit{stability} in the adapter updates using a subspace-aware regularization strategy. Rather than enforcing strict orthogonality between LoRA updates, which can hinder forward transfer, ELLA penalizes overlap with past task-specific directions in proportion to their accumulated energy, thereby discouraging harmful interference while permitting reuse of low-magnitude directions 
that facilitate knowledge transfer, as illustrated in Fig.~\ref{fig:method}. 

Let the LoRA update for task $t$ be $\Delta W_t = A_t B_t$, where $A_t \in \mathbb{R}^{d \times r}$ and $B_t \in \mathbb{R}^{r \times k}$. We construct a cumulative signal from the sum of LoRA-induced weight changes of past tasks:
\vspace{-8pt}
\begin{equation}
\mathcal{W}_{\text{past}} = \sum_{i=1}^{t-1} \Delta W_i.
\vspace{-8pt}
\end{equation}
By construction, $\mathcal{W}_{\text{past}}$ encodes the dominant directions heavily used by prior tasks. 
Motivated by the observation that high-magnitude LoRA components are typically more task-discriminative~\cite{aghajanyan2020intrinsic}, 
we introduce the ELLA alignment penalty as follows:
\vspace{-6pt}
\begin{equation}
    \mathcal{L}_{\text{ELLA}} = \|\Delta W_t * \mathcal{W}_{\text{past}}\|_F^2,
\end{equation}
where $\|\cdot\|_F$ denotes the Frobenius norm. The overall training loss for task $t$ is then:
\vspace{-5pt}
\begin{equation}
    \mathcal{L} = \sum_{(x, y) \in \mathcal{X}_t, \mathcal{Y}_t} \log p_\Theta(y \mid x) + \lambda \cdot \mathcal{L}_{\text{ELLA}}.
\vspace{-5pt}
\label{eq:totalloss}
\end{equation}
Here, $\lambda \geq 0$ controls the trade-off between plasticity (learning the new gradient step) and stability (suppressing interference with past subspaces). 

\paragraph{Formal Characterization:} 
We next formalize the effect of this regularizer, as summarized in the following proposition. The detailed proof is discussed in Appendix~\ref{app:proof}. 

\begin{proposition}
Let $G$ denote the unconstrained gradient step for task $t$, and let 
$E_{ij} = |(\mathcal{W}_{\text{past}})_{ij}| + \varepsilon$ denote the accumulated 
energy of past updates. Then the optimal update $\Delta W^\star_t$ under the ELLA-regularized objective is 
\vspace{-9pt}
\begin{equation}
(\Delta W^\star_t)_{ij} = \frac{G_{ij}}{1 + \lambda E_{ij}^2}
\label{prop1}
\vspace{-9pt}
\end{equation}
Moreover, the interference with past tasks is bounded by
\vspace{-9pt}
\begin{equation}
\bigl|\langle \Delta W^\star_t, \mathcal{W}_{\text{past}}\rangle_F\bigr|
\leq \frac{\|G\|_F}{2\sqrt{\lambda}} \;\|E^{-1}\odot \mathcal{W}_{\text{past}}\|_F
\label{prop2}
\vspace{-4pt}
\end{equation}
\end{proposition}

At a high level, Eq.~\ref{prop1} shows that ELLA acts as an \emph{anisotropic shrinkage operator}: 
coordinates with large past energy are shrunk more aggressively 
while low-energy coordinates remain flexible. The shrinkage strength is directly controlled by $\lambda$: larger values amplify the penalty on high-energy directions, prioritizing stability, while smaller values reduce the penalty, promoting plasticity. This property is further assessed in Sec.~\ref{sec:ablation} and Fig.~\ref{fig:lambda}.
The second result, Eq.~\ref{prop2}, establishes a provable upper bound on task interference, ensuring that 
dominant past directions contribute minimally to forgetting. These properties 
formally justify ELLA’s ability to suppress destructive overlap while reusing 
underutilized subspaces for forward transfer.


\section{Experiments}
\subsection{Datasets} 
\textbf{Standard CL Benchmark (SC)} is a popular CL benchmark designed for language models, comprising five text classification datasets introduced by~\cite{zhang2015character}. Following~\cite{wang2023orthogonal}, we select AG News, Amazon Reviews, DBpedia, and Yahoo Answers, and form three shuffled task sequences, denoted as Orders $1$, $2$, and $3$. 

\textbf{Long Sequence Benchmark (LS)} extends the standard CL setup by incorporating $15$ tasks, including five classification datasets, nine GLUE and SuperGLUE benchmarks, and the IMDB dataset~\cite{razdaibiedina2023progressive}. Following the protocol of prior works~\cite{liao2025data}, we train each task using $1,000$ randomly selected samples, with $500$ samples per class reserved for testing. These tasks are also arranged into shuffled sequences Orders $4$, $5$, and $6$. 

\textbf{TRACE} is a benchmark tailored for CL in LLMs, covering a diverse set of $8$ tasks that span domains such as multiple-choice question answering, multilingual understanding, code generation, and mathematical reasoning~\cite{wang2023trace}. 
See App.~\ref{task:sequence:order} for details on tasks and orderings.

\subsection{Metrics}
Let $a_{i,j}$ denote the testing performance on the $j$-th task after training on the $i$-th task. We evaluate across:
\textbf{Overall Accuracy (OA)}~\cite{chaudhry2018riemannian}: The average accuracy across all tasks after training on the last task, i.e., $\text{OA}_\mathcal{T} = \frac{1}{\mathcal{T}} \sum_{t=1}^{\mathcal{T}} a_{\mathcal{T},t}$; \textbf{Forward Transfer (FWT)}~\cite{lopez2017gradient}: measures how much knowledge from previous tasks transfers to a new task, i.e., $\text{FWT}_\mathcal{T} = \frac{1}{\mathcal{T}} \sum_{t=1}^{\mathcal{T}} (a_{t,t} - a_{0,t})$, where $a_{0,t}$ refers to the performance of training task $t$ individually; \textbf{Backward Transfer (BWT)}~\cite{ke2022continual}: measures how much the learning of subsequent tasks influences the performance of prior tasks, i.e., $\text{BWT}_\mathcal{T} = \frac{1}{\mathcal{T} - 1} \sum_{t=1}^{\mathcal{T} - 1} (a_{\mathcal{T},t} - a_{t,t})$. Moreover, we also report \emph{general ability} (GA) -- the average generalization performance across unseen datasets post CL, 
and \emph{delta general ability} (DeltaGA) -- change in GA relative to the original LLM.

\subsection{Baselines} We compare ELLA against a comprehensive suite of CL baselines. This includes naïve sequential fine-tuning approaches (SeqFT, SeqLoRA), classic regularization methods (EWC~\cite{kirkpatrick2017overcoming}, LwF~\cite{li2017learning}), and modern prompt-tuning techniques (L2P~\cite{wang2022learning}, LFPT5~\cite{qin2021lfpt5}). We also benchmark against state-of-the-art adapter-based methods, including those that enforce strict orthogonality (O-LoRA~\cite{wang2023orthogonal}, LB-CL~\cite{qiao2024learn}) as well as those that combine orthogonality with complex decomposition or replay mechanisms (DATA~\cite{liao2025data}, Recurrent KIF~\cite{feng2025recurrent}). 
Details of baselines are provided in Appendix~\ref{app:baselines}.
\begin{table*}[!ht]
\centering
\resizebox{\textwidth}{!}{
\begin{tabular}{clccccccccc}
\toprule
& \textbf{Methods} & \multicolumn{4}{c}{\textbf{\textcolor{datablue}{{Standard CL Benchmark (SC)}}}} & \multicolumn{4}{c}{\textbf{\textcolor{datablue}{Long Sequence Benchmark (LS)}}} & \textbf{\textcolor{datablue}{TRACE}} \\
& & Order 1 & Order 2 & Order 3 & \textcolor{datablue}{OA} & Order 4 & Order 5 & Order 6 & \textcolor{datablue}{OA} & Order 7 \textcolor{datablue}{OA}\\
\midrule
\multirow{19}{*}{\rotatebox{90}{\textbf{T5-Large}}} 
& SeqFT~\cite{de2019episodic}   & 18.9 & 24.9 & 41.7 & 28.5    & 7.4  & 7.3  & 7.4  & 7.4 & - \\
& SeqLoRA & 39.5 & 31.9 & 46.6 & 39.3   & 4.9  & 3.5  & 4.2  & 4.2      & 12.1 \\
& IncLoRA & 63.4 & 62.2 & 65.1 & 63.6   & 63.0 & 57.9 & 60.4 & 60.5 & - \\
& SeqSVD  & 40.0 & 63.3 & 44.9 & 49.4    & 13.7 & 13.8 & 12.2 & 13.2 & - \\
& EWC~\cite{kirkpatrick2017overcoming}     & 46.3 & 45.3 & 52.1 & 47.9    & 44.9 & 44.0 & 45.4 & 44.8 & - \\
& LwF~\cite{li2017learning}     & 52.7 & 52.9 & 48.4 & 51.3   & 49.7 & 42.8 & 46.9 & 46.5 & - \\
& L2P~\cite{wang2022learning}   & 59.0  &   60.5  &  59.9   & 59.8  & 57.7 & 53.6 & 56.6 & 56.0 &   - \\
& L-CL    & 75.3 & 73.5 & 71.9 & 73.6   & 66.5 & 64.0 & 69.0 & 66.5 & - \\
& B-CL    & 76.4 & 71.5 & 75.1 & 74.3   & 65.7 & 66.4 & 69.2 & 67.1 & - \\
& IncSVD  & 76.0 & 73.4 & 74.0 & 74.5   & 67.6 & 65.3 & 62.6 & 65.2 & - \\
& LB-CL~\cite{qiao2024learn}   & 76.9 & 76.5 & \underline{76.8} & 76.7   & 68.4 & 67.3 & 71.8 & 69.2 & - \\
& O-LoRA~\cite{wang2023orthogonal} & 73.5 & 71.4 & 70.0 & 71.6 & 65.4 & 65.2 & 65.2 & 65.3 & 23.1 \\
& \quad + MIGU~\cite{du2024unlocking} & \underline{77.1} &\underline{77.0}& 75.6 &76.6 & 67.3 &68.5 &74.0 &70.0 & - \\
& DATA~\cite{liao2025data}    &  71.5 & 70.5 & 68.0 & 70.0    & 71.5 & 70.5 & 68.0 & 70.0     & 16.7 \\
& \grayrow{\quad + Replay}  & \grayrow{77.0} & \grayrow{75.6} & \grayrow{75.2} & \grayrow{75.9}
& \grayrow{\textbf{75.6}} & \grayrow{\textbf{73.2}} & \grayrow{\underline{74.1}} & \grayrow{\underline{74.3}} & \grayrow{\underline{36.5}} \\

& \grayrow{LFPT5}~\cite{qin2021lfpt5} & \grayrow{66.6} & \grayrow{71.2} & \grayrow{76.2} & \grayrow{71.3}
& \grayrow{69.8} & \grayrow{67.2} & \grayrow{69.2} & \grayrow{68.7} & \grayrow{-} \\

& \grayrow{SeqLoRAReplay} & \grayrow{4.0} & \grayrow{73.1} & \grayrow{73.0} & \grayrow{73.3}
& \grayrow{\underline{74.2}} & \grayrow{\underline{72.7}} & \grayrow{73.9} & \grayrow{73.6} & \grayrow{34.0} \\

& \grayrow{Recurrent-KIF}~\cite{feng2025recurrent} & \grayrow{-} & \grayrow{-} & \grayrow{-} & \grayrow{\underline{78.4}}
& \grayrow{-} & \grayrow{-} & \grayrow{-} & \grayrow{\textbf{77.8}} & \grayrow{-} \\

& \bluerow{\textbf{ELLA (ours)}} 
& \bluerow{\textbf{80.0}} & \bluerow{\textbf{80.0}} & \bluerow{\textbf{79.8}} & \bluerow{\textbf{79.9}} 
& \bluerow{73.4} & \bluerow{72.0} & \bluerow{\textbf{75.4}} & \bluerow{73.6} & \bluerow{\textbf{40.0}} \\

\midrule
\multirow{6}{*}{\rotatebox{90}{\textbf{LLaMA3.1-8B}}}
& SeqLoRA & 75.88 & 74.40 & 74.35 & 74.86 & 67.81 & 65.93 & 63.80 & 65.85 & 28.23 \\
& O-LoRA~\cite{wang2023orthogonal} & 69.39 & 67.58 & 71.44 & 69.46 & 69.54 & 64.42 & 66.50 & 66.82 & 28.45 \\
& DATA$^\star$~\cite{liao2025data} & 76.10 & 75.69 & 75.52 & 75.77 & 71.34 & 70.77 & 72.95 & 71.69 & 31.33 \\
& \quad + \grayrow{Replay$^\star$} & \grayrow{\underline{77.56}} & \grayrow{\underline{77.01}} & \grayrow{76.03} & \grayrow{\underline{76.83}}
& \grayrow{\textbf{73.10}} & \grayrow{\textbf{73.05}} & \grayrow{74.17} & \grayrow{73.44} & \grayrow{\textbf{34.16}} \\

& \grayrow{SeqLoRAReplay} & \grayrow{77.02} & \grayrow{76.53} & \grayrow{\underline{76.19}} & \grayrow{76.58}
& \grayrow{72.03} & \grayrow{\underline{72.96}} & \grayrow{\underline{75.38}} & \grayrow{\underline{73.46}} & \grayrow{33.02} \\

& \bluerow{\textbf{ELLA (ours)}} 
& \bluerow{\textbf{77.80}} & \bluerow{\textbf{77.20}} & \bluerow{\textbf{77.70}} & \bluerow{\textbf{77.57}} 
& \bluerow{\underline{72.87}} & \bluerow{72.84} & \bluerow{\textbf{76.82}} & \bluerow{\textbf{74.18}} & \bluerow{\underline{33.29}} \\
\bottomrule
\end{tabular}
}
\vspace{-8pt}
\caption{Overall Average Accuracy (OA) comparison of baselines and ELLA (ours) on Standard CL benchmark (Order $1$, $2$, $3$) and Long Sequence benchmark (Order $4$, $5$, $6$) and TRACE (Order $7$) across multiple transfer orders. Methods in \textcolor[gray]{0.55}{gray} \color{black}{rely on replay mechanisms to boost performance.} Best results in \textbf{bold} and second best \underline{underlined}. $\star$ denotes methods that encounter OOM on GPUs smaller than $2 \times$ H$100$s, underscoring their limited scalability.}
\label{tab:results}
\vspace{-6pt}
\end{table*}
\subsection{Implementation Details}
We evaluate ELLA on a range of LLMs including encoder-decoder T$5$ models~\cite{raffel2020exploring} (T$5$-Large, T$5$-XL), and decoder-only LLaMA
models~\cite{touvron2023llama} (LLaMA-$3.1$ $8$B). ELLA training is performed with the DeepSpeed library using V$100$ 16GB GPUs for T5 and A$40$ 48GB GPUs for LLaMA, while larger baselines such as DATA~\cite{liao2025data} require H$100$ 80GB GPUs. All experiments are
performed with instruction tuning~\cite{wei2021finetuned}. 
Following~\cite{wang2023orthogonal}, we apply LoRA modules (rank=$8$) to the Attention Q-V layers and report the average result of $3$ runs. More details in App.~\ref{app:implementation_details}.

\subsection{Results}
Table~\ref{tab:results} reports overall performance across three CL benchmarks. Across all architectures, task orders, and evaluation settings, \textbf{ELLA consistently sets a new state of the art in replay-free CL for LLMs}. Remarkably, it not only dominates replay-free baselines but also surpasses strong replay-augmented methods, despite introducing no additional memory or computational burden.

On the Standard CL benchmark, ELLA achieves an average accuracy of $79.9$ on T$5$-Large, outperforming the strongest replay-free competitor~\cite{qiao2024learn} and exceeding DATA~\cite{liao2025data} and Recurrent KIF~\cite{feng2025recurrent}, both of which depend on data rehearsal. On the more demanding Long Sequence (LS) setup, ELLA improves over~\cite{qiao2024learn, du2024unlocking} by upto $4.3\%$, demonstrating robustness under long horizons where interference typically compounds. This is further examined in Sec.~\ref{sec:ablation} and Fig.~\ref{fig:change}. On TRACE, which stresses cross-domain CL adaptation, ELLA achieves a dramatic gain of up to $+23.3$, showing its ability to generalize to reasoning, code, and multilingual tasks.

\begin{figure}[t]
\centering
\includegraphics[width=0.5\textwidth]{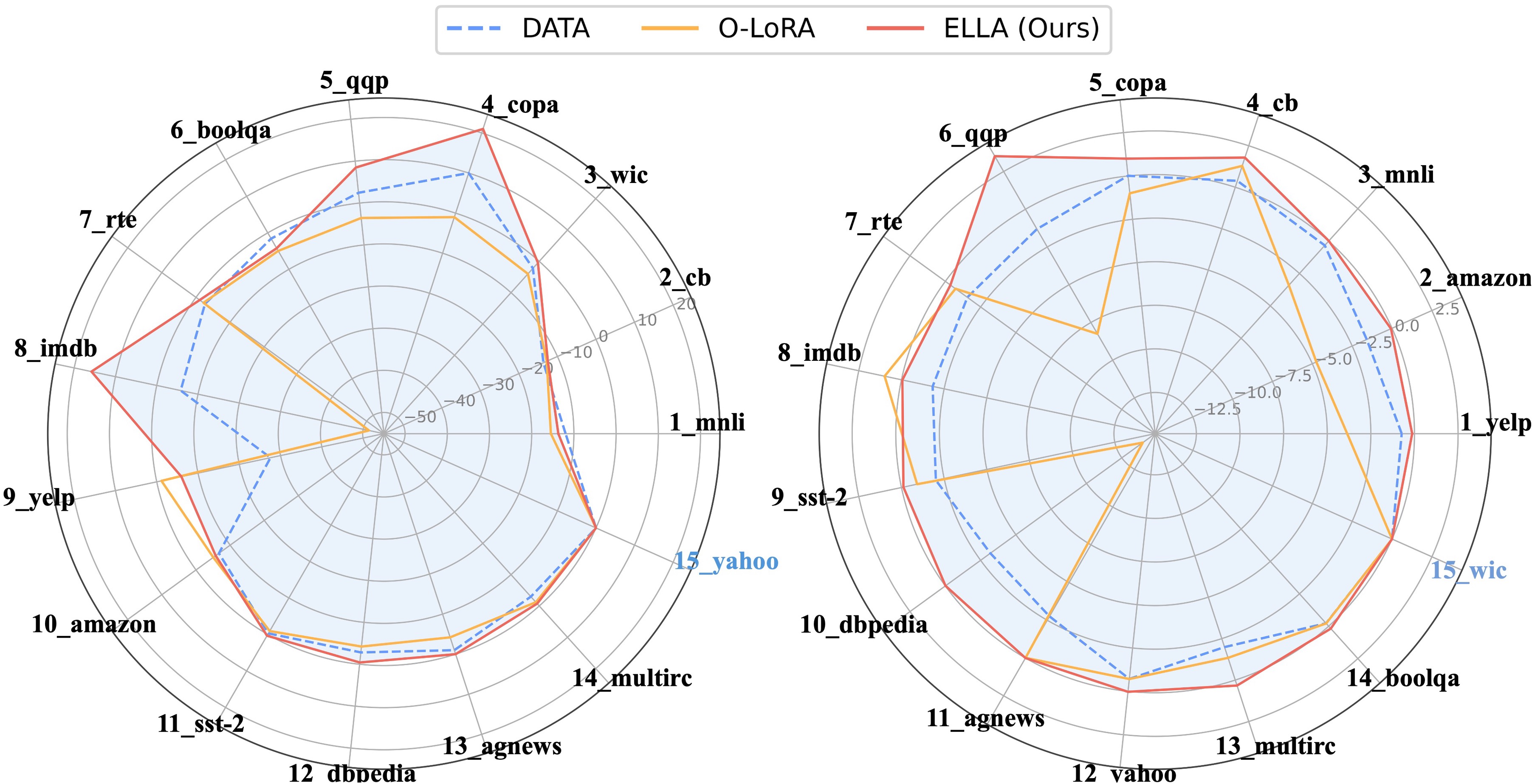} 
\vspace{-19pt}
\caption{Performance impact on Order $4$ (left) and $6$ (right) in terms of BWT. We demonstrate superior resistance to performance decline than baselines (higher values indicate better retention of prior task performance).}
\label{fig:bwt}
\end{figure}

\begin{figure}[t]
\centering
\includegraphics[width=0.5\textwidth]{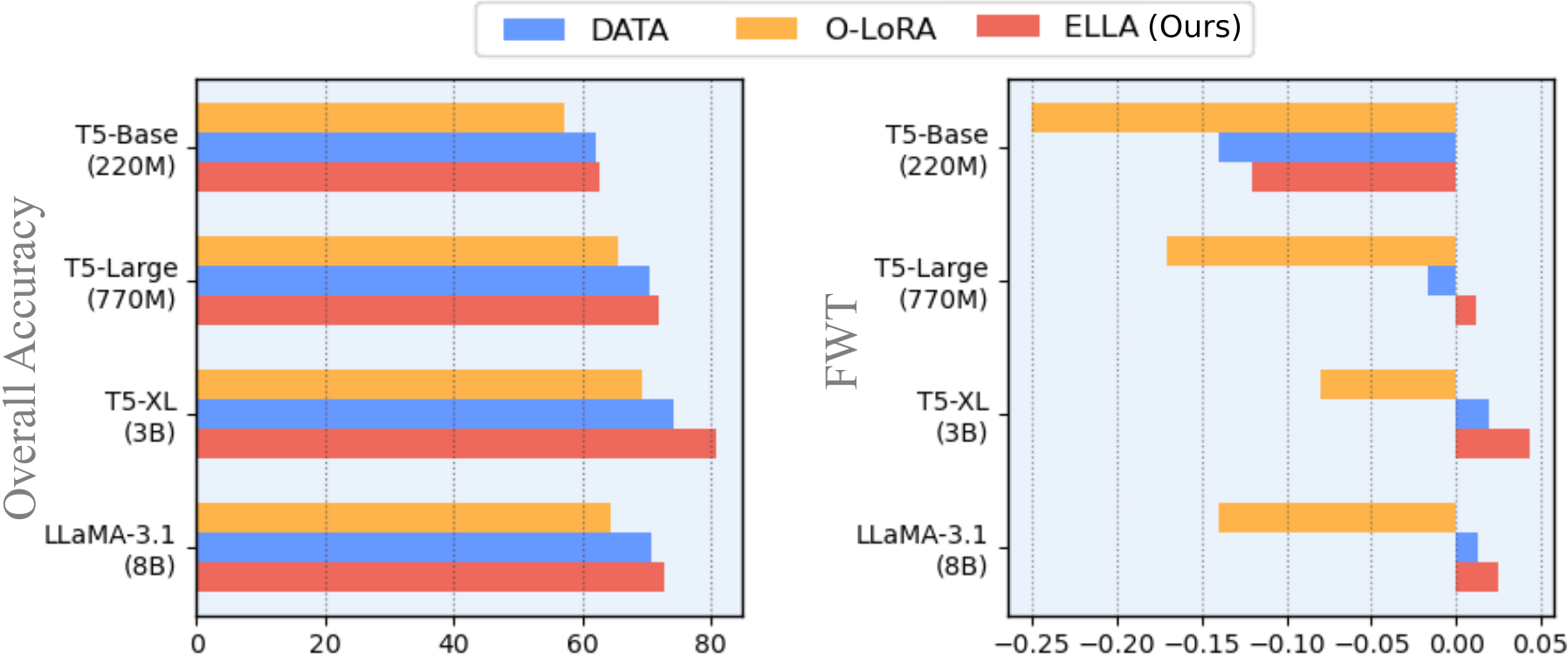} 
\vspace{-20pt}
\caption{Performance comparison across different backbone size and model families. 
}
\label{fig:fwt}
\end{figure}

Beyond average accuracy, Fig.~\ref{fig:bwt} and~\ref{fig:fwt} reveal deeper dynamics. ELLA achieves the best BWT, sharply reducing forgetting, while also obtaining the highest FWT, signaling effective knowledge reuse across tasks. This is crucial: unlike orthogonality-based approaches~\cite{wang2023orthogonal, qiao2024learn}, which guarantee stability but block positive transfer, ELLA explicitly allows low-energy subspace reuse, yielding consistent forward gains. Conversely, replay-based approaches achieve modest transfer but at the cost of large buffers and compute. Notably, ELLA sustains high performance on interference-sensitive tasks like QQP, IMDB, and DBpedia, where baselines degrade sharply. ELLA thus demonstrates that selective subspace decorrelation provides a more principled way to reconcile stability and plasticity than existing strategies. 

Equally important, ELLA achieves these gains with \emph{no replay buffer, no generative augmentation, and no task identity signals}. Replay-heavy methods~\cite{liao2025data, feng2025recurrent} not only incur large memory overheads but also encounter out-of-memory failures on LLaMA-3.1 8B when trained on GPUs smaller than H$100$s. ELLA avoids these pitfalls entirely, remaining lightweight, scalable, and stable even at billion-parameter scale.

In summary, ELLA combines state-of-the-art replay-free accuracy with superior forward transfer, minimal forgetting, and low overhead, setting a new benchmark for scalable CL in LLMs.


\subsection{Task Generalization} 
LLMs that are continually trained on new tasks often exhibit a decline in general performance, revealing catastrophic forgetting of their original capabilities. To assess this aspect, we evaluate the generalization ability of ELLA in the context of continual learning, focusing on its performance on five unseen cross-task benchmarks: MMLU~\cite{hendrycks2020measuring}, GSM$8$k~\cite{cobbe2021training}, BBH~\cite{suzgun2022challenging}, AGIEval~\cite{zhong2023agieval}, and PIQA~\cite{bisk2020piqa}. More details on these datasets are provided in Appendix~\ref{app_expsettings}. 

We begin with a fine-tuned LLaMA $3.1$-$8$B model trained on Order~1 of the SC Benchmark. As shown in Table~\ref{tab:ga_eval}, ELLA consistently achieves higher accuracy across all benchmarks compared to baselines, demonstrating its effectiveness in preserving knowledge while enabling strong cross-task generalization. Crucially, we see improved performance even over the original zero-shot model, underscoring its superiority in effectively combining acquired knowledge to address novel tasks. Beyond mitigating forgetting, ELLA delivers positive \textit{DeltaGA}, showing that it not only retains prior skills but also facilitates forward transfer to unseen tasks. This indicates that ELLA provides a stable, rehearsal-free path to cross-task generalization, making it valuable for the continual deployment of LLMs in dynamic environments.

\begin{table}[t]
\centering
\resizebox{\columnwidth}{!}{
\begin{tabular}{lcccccc}
\toprule
\textbf{Methods} & \textbf{\textcolor{datablue}{MMLU}} & \textbf{\textcolor{datablue}{GSM8K}} & \textbf{\textcolor{datablue}{BBH}} & \textbf{\textcolor{datablue}{PIQA}} & \textbf{\textcolor{datablue}{AGIEval}} & \textbf{DeltaGA} \\
\midrule
Zero Shot        & 59.64 & 69.67 & 35.65 & 72.42 & 43.92 & 0 \\
\hdashline
SeqLoRA          & 56.05 & 70.20 & 35.16 & 67.85 & 41.02 & -11.02 \\
O-LoRA           & 57.36 & 66.79 & 33.82 & 64.31 & \textbf{44.21} & -14.81 \\
\rowcolor{gray!25} SeqLoRAReplay & 55.81 & 66.43 & 32.90 & 67.77 & 41.58 & -16.81 \\
\rowcolor{gray!25} DATA & 46.84& 14.17& 12.02& 59.02& 39.11&-110.14\\
\rowcolor{tabblue!15} \textbf{Ours}    & \textbf{59.92} & \textbf{71.05} & \textbf{36.37} & \textbf{75.52} & 43.92 & \textbf{5.48} \\
\bottomrule
\end{tabular}
}
\vspace{-7pt}
\caption{Generalization ability (GA) on unseen tasks. 
}
\vspace{-9pt}
\label{tab:ga_eval}
\end{table}
\subsection{Scale to Larger Models}
To assess the scalability and robustness of ELLA across model sizes and architectures, we examine CL performance across T$5$-Base ($220$M), T$5$-Large ($770$M), T$5$-XL ($3$B), and LLaMA-$3.1$ $8$B on Order $5$ of the Long Sequence Benchmark. As seen in Fig.~\ref{fig:fwt}, ELLA demonstrates consistent gains in average performance and transfer metrics as model size increases, outperforming baselines across all backbone configurations. 

It is worth noting that even with the largest backbone model, O-LoRA($8$B) still falls short in terms of FWT compared to the smallest version of ELLA ($220$M). This further highlights the crucial importance of selecting the pertinent PEFT algorithm for continual adaptation, rather than relying solely on scaling backbone size. Notably, the T$5$-XL model achieves higher OA than the larger LLaMA-$3.1$ model, highlighting that encoder-decoder architectures like T$5$ might be more effective for CL on classification tasks than decoder-only LLMs. We attribute this to the full-sequence bidirectional attention and cross-attention present in encoder-decoder designs, which better facilitate knowledge retention and task transfer than the causally masked architecture of models like LLaMA. These findings underscore that, beyond scaling size, architectural choice plays a crucial role, and shows that forgetting is model-dependent. 

\begin{table}[t]
\centering
\resizebox{\columnwidth}{!}{
\begin{tabular}{lcccc}
\toprule
\textbf{Method} & \textbf{Trainable Params} & \textbf{Storage (MB)} & \textbf{Replay} & \textbf{Time/Epoch (mins)} \\
\midrule
SeqLoRA & 0.062 & 0 & 0 & 4 \\
O-LoRA & 0.062 & 31.46 & 0 & 4.5 \\
\rowcolor{gray!25} SeqLoRAReplay & 0.062 & 0 & 2\% & 4 \\
\rowcolor{gray!25} DATA & 0.369 & 147.46 & 2\% & 6.5 \\
\rowcolor{tabblue!15} \textbf{Ours} & 0.062 & 4.19 & 0 & 4.5 \\
\bottomrule
\end{tabular}}
\vspace{-9pt}
\caption{Comparison of training efficiency, memory overhead, and storage requirements with T$5$-Large.}
\label{tab:efficiency}
\vspace{-7pt}
\end{table}
\begin{figure}[t]
\centering
\includegraphics[width=0.45\textwidth]{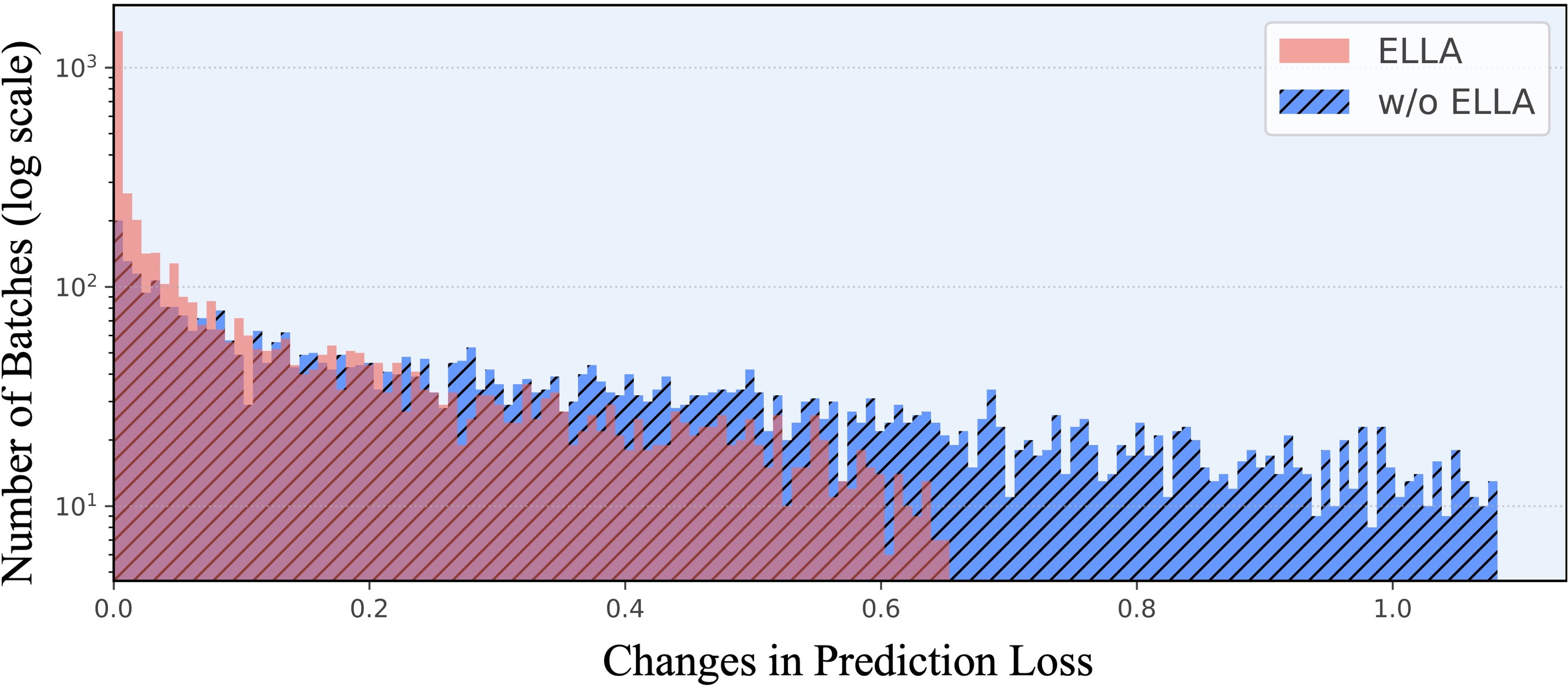} 
\vspace{-9pt}
\caption{Histogram of prediction loss changes after training on a new task. The \textcolor{tabred!80}{ELLA constraint} helps reduce the changes -- preserve the loss of previous
tasks -- in comparison to when it is not present (\textcolor{tabblue}{$\lambda = 0$}).}
\label{fig:loss}
\end{figure}
\subsection{Efficiency Analysis} 
As shown in Table~\ref{tab:efficiency}, ELLA matches O-LoRA in trainable parameters while cutting storage 
from $31.46$MB to just $4.19$MB--an $8\times$ reduction. Unlike replay-based methods such as DATA and SeqLoRAReplay, ELLA requires no buffer or feature storage and adds only minimal runtime cost. By embedding a lightweight regularization penalty into the loss instead of storing multiple task-specific modules, ELLA achieves strong efficiency in memory, compute, and training time -- showing that ELLA is scalable and practical for long-horizon CL, where resource demands are typically prohibitive.
\section{Discussions}
\label{sec:ablation}
\hspace{8pt}\textbf{Does ELLA preserve previous task performance during CL?}
We examine the effectiveness of ELLA in preventing degradation on previously learned tasks by measuring the change in prediction loss on past-task batches after training a new task. As shown in Fig.~\ref{fig:loss}, ELLA significantly reduces the number of batches experiencing large increases in loss, especially in the high-loss tail region. This indicates that ELLA better preserves useful gradients from prior tasks by leveraging its alignment penalty strategy. In contrast, the baseline (without ELLA) exhibits a broader distribution of loss spikes, revealing higher susceptibility to catastrophic forgetting. These results confirm that ELLA enhances stability across tasks without requiring replay or task labels.

\begin{figure}[t]
\centering
\includegraphics[width=0.48\textwidth]{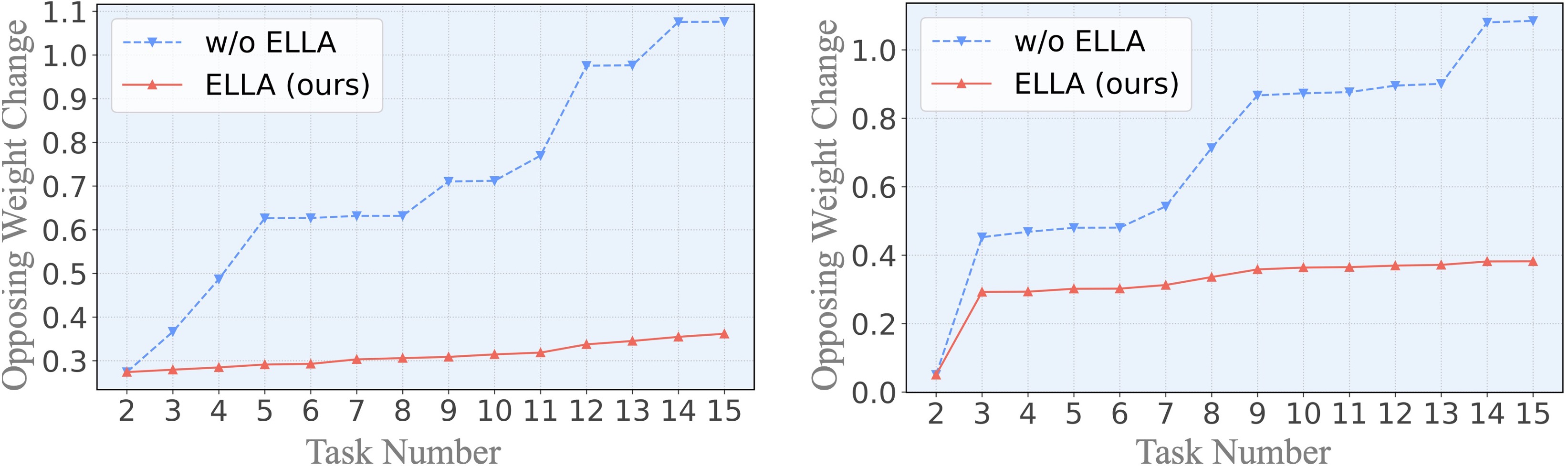} 
\vspace{-22pt}
\caption{Opposing direction weight change across task sequence for T$5$-Large (left) and LLaMA-$3.1$ $8$B (right). ELLA consistently reduces backward-conflicting updates, promoting stable continual adaptation.}
\vspace{-5pt}
\label{fig:change}
\end{figure}

\textbf{Directional Consistency of Updates Over Task Sequence.}
In Fig.~\ref{fig:change}, we analyze the degree of opposing weight updates after each new task for Order $5$ on both T$5$-Large and LLaMA$3.1$-$8$B. Specifically, we measure the magnitude of weight change that occurs in the direction opposite to that of the previous task, indicating a misalignment between updates over time. Standard LoRA exhibits consistently high opposing updates, suggesting that new learning often disrupts previously acquired representations, while ELLA significantly reduces such opposing direction weight updates, enabling smoother and more stable knowledge accumulation across tasks. This improvement is consistent across both encoder-decoder and decoder-only models.

\begin{figure}[t]
\centering
\includegraphics[width=0.45\textwidth]{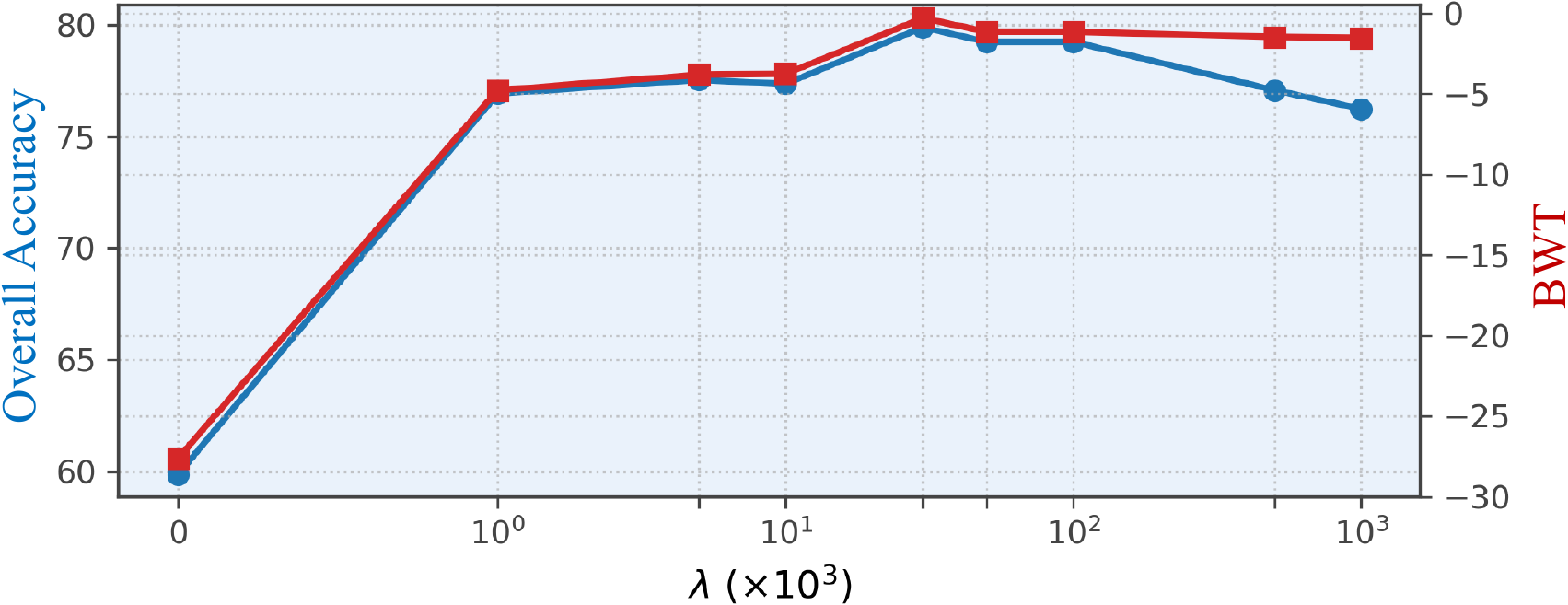} 
\vspace{-8pt}
\caption{Impact of $\lambda$ scaling on Order $1$. \textcolor{tabblue}{Overall Accuracy} and \textcolor{tabred!80}{BWT} as $\lambda$ varies 
shows that moderate scaling achieves the best stability-plasticity balance.}
\vspace{-3pt}
\label{fig:lambda}
\end{figure}

\textbf{Impact of $\lambda$ Scaling.} 
Fig.~\ref{fig:lambda} shows that $\lambda=0$ (i.e, a vanilla LoRA) results in severe forgetting, as evidenced by low OA and BWT, while increasing $\lambda$ progressively improves both metrics by constraining interference. Excessively large values, however, hinder new task adaptation, lowering OA. The best performance arises at moderate $\lambda$, where the two terms in Eq.~\ref{eq:totalloss} are balanced, yielding both stability and plasticity. This behavior is consistent with our theoretical characterization in Appendix~\ref{app:proof}, which shows that $\lambda$ acts as an anisotropic shrinkage factor in our update, interpolating between unconstrained learning and stability-preserving adaptation.
\begin{table}[t]
\centering
\resizebox{0.8\columnwidth}{!}{
\begin{tabular}{lcccc}
\toprule
\textbf{LoRA\_dim} & Order 1 & Order 2 & Order 3 & \textbf{Avg} \\
\midrule
2  & 72.29 & 74.00 & 77.08 & 74.46 \\
4  & 73.22 & 75.15 & 77.72 & 75.36 \\
\rowcolor{tabblue!15} \textbf{8}  & \textbf{79.95} & \textbf{80.00} & \textbf{79.82} & \textbf{79.92} \\
16 & 77.38 & 77.65 & 76.19 & 77.07 \\
\bottomrule
\end{tabular}
}
\vspace{-7pt}
\caption{Impact of LoRA rank on SC Benchmark.}
\vspace{-7pt}
\label{tab:lora_dim}
\end{table}

\textbf{Studying Optimal LoRA Rank for Plasticity-Stability Tradeoff.} We investigate how LoRA rank influences CL by evaluating ELLA with T$5$-Large across three task orders while varying the LoRA dimension \(r\). As shown in Table~\ref{tab:lora_dim}, performance improves with increasing \( r \), peaking at \( r=8 \). Very low ranks (\(r=2\)) severely limit plasticity and hinder adaptation to new tasks, while higher ranks (\( r=16 \)) reduce stability by overfitting to current tasks. This underscores that careful rank selection is 
crucial for sustaining continual adaptation without catastrophic forgetting.

\section{Conclusion}
We introduced \textbf{ELLA}, a simple yet powerful approach for continual adaptation of LLMs that mitigates 
forgetting without replay or task identifiers. By encouraging controlled de-alignment from past subspaces, 
ELLA reduces destructive interference while enabling reuse of underutilized directions. We provide formal 
guarantees showing that ELLA’s update admits a closed-form solution with bounded interference, offering a principled foundation for balancing plasticity and stability. Extensive experiments across benchmarks, model architectures, and scales demonstrate that ELLA consistently improves both forward and backward transfer, while remaining highly parameter- and memory-efficient. Notably, ELLA surpasses state-of-the-art baselines and scales robustly to larger backbones such as T5-$3$B and LLaMA-$3.1$-$8$B with accuracy
gains of up to $9.6\%$. It also uniquely enhances generalization performance on previously unseen tasks compared to the original LLM models -- an ability absent in prior methods. These findings position ELLA as a lightweight, scalable, and universal method for lifelong adaptation in LLMs, demonstrating that efficiency and continual improvement need not come at the cost of forgetting.

\section{Limitations} While our approach demonstrates strong performance, its scalability to more complex continual learning scenarios involving hundreds of tasks remains an open question. Moreover, although task identification is not required at inference time, our current training still assumes task labels to assign task-specific LoRA parameters. Developing task-agnostic training strategies is a promising direction for future work. Finally, due to resource constraints, we have not evaluated our method on larger models like LLaMA-$3.1$-$70$B, and extending ELLA to continual multimodal LLMs remains an exciting avenue for exploration.


\section{Ethical Considerations}
ELLA represents a significant step towards practical and scalable lifelong learning for LLMs. By enabling efficient, replay-free updates, our work makes it more feasible to keep models current in dynamic environments without hindering data privacy and reducing the environmental footprint of AI systems. While ELLA's design promotes efficiency and privacy, it does not inherently mitigate the risks of misuse, such as the generation of biased or harmful content, which are endemic to the base models. Deploying models updated with ELLA requires a continued commitment to responsible data curation, fairness audits, and robust safety mechanisms to address these challenges. The datasets and models we used are public, hence there are no privacy issues.

\section{AI Writing Statement}
This paper utilized AI assistance for language polishing of the manuscript, including vocabulary correction and spell checking.
\bibliography{acl_main}
\appendix
\section*{Appendix}

This appendix provides supplementary material to support the findings presented in the main paper. We begin in Section~\ref{app:proof} with a detailed theoretical analysis of the ELLA regularizer, offering a formal proof of its properties, including its formulation as an anisotropic shrinkage operator and its ability to bound interference provably. In Section~\ref{app_expsettings}, we provide a comprehensive overview of our experimental settings. This includes detailed descriptions of the datasets, baselines, task sequences, instruction prompts, and specific hyperparameter configurations used in our evaluations. Finally, Section~\ref{license} lists and discusses the artifact licenses.
\section{Theoretical Analysis of the
ELLA Regularizer}
\label{app:proof}
Our theoretical analysis shows that the ELLA regularizer has two key properties: (i) its optimal update is an anisotropic shrinkage operator that selectively preserves plasticity, and (ii) it establishes a provable bound on the interference with past tasks, guaranteeing stability.

\subsection*{A.1 Setup and Preliminaries}
Our analysis is set in the context of continual learning, where a model adapts to a sequence of tasks.
For each new task $t$, the model is adapted by learning a low-rank update matrix, $\Delta W_t$, using the LoRA method. This update is defined as the product of two smaller matrices:
\begin{equation}
\Delta W_t = A_t B_t
\end{equation}
where $A_t\in\mathbb{R}^{d \times r} \text{ and }B_t \in \mathbb{R}^{r \times k} \text{ with rank } r \ll \min\{d,k\}$. This matrix represents the specific knowledge acquired for task $t$. We define the aggregated matrix of past updates as:
\begin{equation}
\mathcal{W}_{past} = \sum_{i=1}^{t-1} \Delta W_i
\end{equation}

The practical regularization term used in our work is $\|\mathcal{W}_{past} \odot \Delta W_t\|_F^2$. For analytical tractability, we replace this term with a closely related objective. We introduce an energy matrix $E \in \mathbb{R}^{d \times k}$ derived from past updates, be defined as $E_{ij} = |(\mathcal{W}_{past})_{ij}| + \varepsilon$ for a small constant $\varepsilon > 0$ that ensures all entries are positive and $E^{-1}$ is well-defined. This preserves the core principle of suppressing alignment with high-energy coordinates.

For the current task $t$, there is an ideal, unconstrained update that would best minimize the task loss, $\mathcal{L}_t$. This is the standard gradient step, which we denote as $G$.
\[
G \triangleq \nabla \mathcal{L}_t(W_{t-1})
\]
We think of $G$ as the update we would make if we did not have to worry about forgetting past knowledge.

The goal of ELLA is to find an optimal update $\Delta W_t$ that is close to the target gradient $G$ (learning the new task) but is penalized for changing parameters that were important for past tasks (not forgetting). This is formulated as the following regularized optimization problem:
\begin{equation}
\label{eq:ella-prob}
\min_{\Delta W}\;\; \underbrace{\frac{1}{2}\|\Delta W_t - G\|_F^2}_{\text{Plasticity Term}} \;+\; \underbrace{\frac{\lambda}{2}\,\|E\odot \Delta W_t\|_F^2}_{\text{Stability Term}}
\end{equation}
where $\lambda > 0$ is the regularization strength and $\odot$ denotes the element-wise Hadamard product. The \textit{plasticity term} encourages $\Delta W$ to be similar to $G$, while the \textit{stability term} penalizes changes to coordinates with high energy (large values in $E$).


\textbf{Proposition 1}
The solution $\Delta W_t^\star$ to the ELLA objective in Eq. \eqref{eq:ella-prob} has the following properties:
\begin{enumerate}
    \item[\textit{(i)}] It is an anisotropic shrinkage operator applied to the unconstrained step $G$, with the closed-form solution:
    \[
    (\Delta W_t^\star)_{ij} = \frac{G_{ij}}{\,1+\lambda E_{ij}^2\,}
    \]
    \item[\textit{(ii)}] The interference with past updates, measured by the inner product $\langle \Delta W_t^\star, \mathcal{W}_{past} \rangle_F$, is bounded as follows:
    \[
    \bigl|\langle \Delta W_t^\star,\mathcal{W}_{past}\rangle_F\bigr|
    \le
    \frac{\|G\|_F}{2\sqrt{\lambda}}\;\|E^{-1}\odot \mathcal{W}_{past}\|_F
    \]
\end{enumerate}


\begin{proof}
\textbf{Proof of part (i):} The objective function in Eq. \eqref{eq:ella-prob} is strongly convex since its second derivative is strictly positive, and separable across the matrix entries $(i,j)$. We can therefore solve for each entry $(\Delta W_t)_{ij}$ independently. Let $z = (\Delta W_t)_{ij}$, $g = G_{ij}$, and $e = E_{ij}$. The per-coordinate objective is:
\[
\min_{z\in\mathbb{R}} \;\; \tfrac12(z-g)^2 + \tfrac{\lambda}{2} e^2 z^2
\]
Taking the derivative with respect to $z$ and setting it to zero yields the optimal solution $z^\star$:
\[
(z^\star-g) + \lambda e^2 z^\star = 0 \quad \Rightarrow \quad (1+\lambda e^2)\,z^\star = g
\]
This gives the closed-form solution for each coordinate, proving part (i):
\begin{equation}
\label{eq:closed-form}
(\Delta W_t^\star)_{ij} = \frac{G_{ij}}{\,1+\lambda E_{ij}^2\,}
\end{equation}
This operator selectively shrinks the update for each coordinate based on the accumulated energy of past updates $E_{ij}$, with larger energy leading to stronger shrinkage.

\textbf{Proof of part (ii):} Next, we first bound the norm of the energy-weighted update. Using the solution from Eq. \eqref{eq:closed-form}:
\begin{align}
    \|E\odot \Delta W_t^\star\|_F^2
&=
\sum_{ij} \frac{E_{ij}^2\,G_{ij}^2}{(1+\lambda E_{ij}^2)^2} \\
&\le
\left(\sup_{x\ge0} \frac{x}{(1+\lambda x)^2}\right)\, \sum_{ij}G_{ij}^2
\end{align}
The supremum is attained at $x=1/\lambda$, yielding $\sup_{x\ge0} \frac{x}{(1+\lambda x)^2} = \frac{1}{4\lambda}$. Therefore,
\begin{equation}
    \|E\odot \Delta W_t^\star\|_F^2\le\frac{1}{4\lambda}\,\|G\|_F^2
\label{eq:penalty-bound}
\end{equation}
This shows that the effective penalty is never worse than a $\tfrac{1}{4\lambda}$ fraction of the gradient energy, ensuring the shrinkage remains controlled.

Now, we try to bound the interference term $\langle \Delta W_t^\star,\mathcal{W}_{past}\rangle_F$. By rewriting the inner product and applying the Cauchy-Schwarz inequality:
\begin{align}
&\bigl|\langle \Delta W_t^\star,\mathcal{W}_{past}\rangle_F\bigr| \\
&=
\bigl|\langle E\odot \Delta W_t^\star,\, E^{-1}\odot \mathcal{W}_{past}\rangle_F\bigr|\\
&\le
\|E\odot \Delta W_t^\star\|_F\ \|E^{-1}\odot \mathcal{W}_{past}\|_F
\label{eq:cauchy}
\end{align}
Substituting the bound from Eq. \eqref{eq:penalty-bound} into Eq. \eqref{eq:cauchy} yields the final interference bound:
\[
\bigl|\langle \Delta W_t^\star,\mathcal{W}_{past}\rangle_F\bigr|
\le
\frac{\|G\|_F}{2\sqrt{\lambda}}\;\|E^{-1}\odot \mathcal{W}_{past}\|_F
\]
This completes the proof of part (ii). The term $\|E^{-1}\odot \mathcal{W}_{past}\|_F$ ensures that coordinates with high energy (large $E_{ij}$) contribute minimally to the bound, formally proving that ELLA enforces stability by suppressing interference in high-energy coordinates, while preserving plasticity where capacity remains.

\end{proof}

\section{Experimental Settings}
\label{app_expsettings}
\subsection{Datasets}

\hspace{10pt}\textbf{Train Tasks.} Tables~\ref{tab:longseq_datasets} and ~\ref{tab:trace_datasets} provide detailed information on the datasets utilized in our continual
learning (CL) experiments. Table~\ref{tab:longseq_datasets} presents the $15$
datasets included in the Long Sequence Benchmark~\cite{razdaibiedina2023progressive}, while Table~\ref{tab:trace_datasets} outlines
the $8$ datasets from TRACE~\cite{wang2023trace}. Both tables include the corresponding evaluation
metrics for each dataset.

\textbf{Unseen tasks.} We select the following datasets to test generalization performance post continual adaptation: (1) Multitask Language Understanding (MMLU)~\cite{hendrycks2020measuring}, which includes multiple-choice questions across $57$ subjects. (2) GSM8K~\cite{cobbe2021training}, which is a high-quality linguistically diverse multi-step elementary math reasoning dataset. (3) BIG-Bench Hard (BBH)~\cite{suzgun2022challenging},
which includes $27$ challenging tasks spanning arithmetic, symbolic reasoning, and more, derived from
BIG-Bench (BB)~\cite{srivastava2022beyond}. Most of the data consists of multiple-choice questions. (4) AGIEval~\cite{zhong2023agieval}, which includes a wide range of high-quality official entrance exams, qualifying exams, and advanced competitions tailored to human participants. (5) PIQA~\cite{bisk2020piqa} which is a dataset for commonsense reasoning, and was created to investigate the physical knowledge of existing models in NLP.

\begin{table*}[!h]
\centering
\begin{tabular}{lllll}
\toprule
\textbf{Dataset Name} & \textbf{Category} & \textbf{Task} & \textbf{Domain} & \textbf{Metric} \\
\midrule
Yelp       & CL Benchmark & Sentiment Analysis               & Yelp Reviews            & Accuracy \\
Amazon     & CL Benchmark & Sentiment Analysis               & Amazon Reviews          & Accuracy \\
DBPedia    & CL Benchmark & Topic Classification             & Wikipedia               & Accuracy \\
Yahoo      & CL Benchmark & Topic Classification             & Yahoo Q\&A              & Accuracy \\
AG News    & CL Benchmark & Topic Classification             & News                    & Accuracy \\
MNLI       & GLUE         & Natural Language Inference       & Various                 & Accuracy \\
QQP        & GLUE         & Paragraph Detection              & Quora                   & Accuracy \\
RTE        & GLUE         & Natural Language Inference       & News, Wikipedia         & Accuracy \\
SST-2      & GLUE         & Sentiment Analysis               & Movie Reviews           & Accuracy \\
WiC        & SuperGLUE    & Word Sense Disambiguation        & Lexical Databases       & Accuracy \\
CB         & SuperGLUE    & Natural Language Inference       & Various                 & Accuracy \\
COPA       & SuperGLUE    & Question and Answering           & Blogs, Encyclopedia     & Accuracy \\
BoolQA     & SuperGLUE    & Boolean Question and Answering   & Wikipedia               & Accuracy \\
MultiRC    & SuperGLUE    & Question and Answering           & Various                 & Accuracy \\
IMDB       & SuperGLUE    & Sentiment Analysis               & Movie Reviews           & Accuracy \\
\bottomrule
\end{tabular}
\caption{
The details of 15 classification datasets in the Long Sequence Benchmark~\cite{razdaibiedina2023progressive}. The first five tasks correspond to the standard CL benchmark~\cite{zhang2015character}.
}
\label{tab:longseq_datasets}
\end{table*}

\begin{table*}[!h]
\centering
\begin{tabular}{llcccc}
\toprule
\textbf{Dataset}      & \textbf{Source}   & \textbf{Avg len} & \textbf{Metric}         & \textbf{Language} & \textbf{\#Data} \\
\midrule
\multicolumn{6}{l}{\cellcolor{gray!15}\textit{Domain-specific}} \\
ScienceQA   & Science    & 210   & Accuracy    & English & 5,000 \\
FOMC        & Finance    & 51    & Accuracy    & English & 5,000 \\
MeetingBank & Meeting    & 2853  & ROUGE-L     & English & 5,000 \\
\multicolumn{6}{l}{\cellcolor{gray!15}\textit{Multi-lingual}} \\
C-STANCE    & Social media & 127 & Accuracy    & Chinese & 5,000 \\
20Minuten   & News         & 382 & SARI        & German  & 5,000 \\
\multicolumn{6}{l}{\cellcolor{gray!15}\textit{Code Completion}} \\
Py150       & Github    & 422   & Edim Similarity & Python  & 5,000 \\
\multicolumn{6}{l}{\cellcolor{gray!15}\textit{Mathematical Reasoning}} \\
NumGLUE-cm  & Math      & 32    & Accuracy    & English & 5,000 \\
NumGLUE-ds  & Math      & 21    & Accuracy    & English & 5,000 \\
\bottomrule
\end{tabular}
\caption{
The overview of dataset statistics in TRACE~\cite{wang2023trace}. The `Source' indicates the origin of the context. `Avg len' denotes the average length, measured in word count for English, German, and code datasets, and in character count for Chinese datasets. `SARI' is a score that is specific to evaluating simplification tasks.
}
\label{tab:trace_datasets}
\end{table*}

\begin{table*}[h]
\centering
\begin{tabular}{lcp{10cm}}
\toprule
\textbf{Benchmark} & \textbf{Order} & \textbf{Task Sequence} \\
\midrule
\multirow{3}{*}{Standard CL Benchmark} 
  & 1 & dbpedia $\rightarrow$ amazon $\rightarrow$ yahoo $\rightarrow$ ag \\
  & 2 & dbpedia $\rightarrow$ amazon $\rightarrow$ ag $\rightarrow$ yahoo \\
  & 3 & yahoo $\rightarrow$ amazon $\rightarrow$ ag $\rightarrow$ dbpedia \\
\midrule
\multirow{3}{*}{Long Sequence Benchmark} 
  & 4 & mnli $\rightarrow$ cb $\rightarrow$ wic $\rightarrow$ copa $\rightarrow$ qqp $\rightarrow$ boolqa $\rightarrow$ rte $\rightarrow$ imdb $\rightarrow$ yelp $\rightarrow$ amazon $\rightarrow$ sst-2 $\rightarrow$ dbpedia $\rightarrow$ ag $\rightarrow$ multirc $\rightarrow$ yahoo \\
  & 5  & multirc $\rightarrow$ boolqa $\rightarrow$ wic $\rightarrow$ mnli $\rightarrow$ cb $\rightarrow$ copa $\rightarrow$ qqp $\rightarrow$ rte $\rightarrow$ imdb $\rightarrow$ sst-2 $\rightarrow$ dbpedia $\rightarrow$ ag $\rightarrow$ yelp $\rightarrow$ amazon $\rightarrow$ yahoo \\
  & 6 & yelp $\rightarrow$ amazon $\rightarrow$ mnli $\rightarrow$ cb $\rightarrow$ copa $\rightarrow$ qqp $\rightarrow$ rte $\rightarrow$ imdb $\rightarrow$ sst-2 $\rightarrow$ dbpedia $\rightarrow$ ag $\rightarrow$ yahoo $\rightarrow$ multirc $\rightarrow$ boolqa $\rightarrow$ wic \\
\midrule
\multirow{1}{*}{TRACE}
  & 7 & c-stance $\rightarrow$ fomc $\rightarrow$ meetingbank $\rightarrow$ py150 $\rightarrow$ scienceqa $\rightarrow$ numglue-cm $\rightarrow$ numglue-ds $\rightarrow$ 20minuten \\
\bottomrule
\end{tabular}
\caption{
Seven distinct orders of task sequences were employed for the experiments in continual learning. Orders 1-3 align with the Standard CL Benchmarks, as adopted in previous studies~\cite{liao2025data}. Orders 4-6 pertain to the Long Sequence Benchmarks, which encompass a total of 15 tasks~\cite{razdaibiedina2023progressive}. Order 7 refers to the TRACE benchmark, specifically designed for LLMs, and comprises eight datasets~\cite{wang2023trace}.
}
\label{tab:task_orders}
\end{table*}
\subsection{Task Sequence Orders}
\label{task:sequence:order}
We report all task orders used for our CL experiments
in Table~\ref{tab:task_orders}.

\subsection{Baselines}
\label{app:baselines}
We compare our method against a comprehensive set of recent CL baselines, detailed as follows:
(1)~\textbf{SeqFT} sequentially fine-tunes all model parameters without CL mechanisms like regularization or replay (RorR). 
(2)~\textbf{SeqLoRA} applies fixed-size LoRA tuning per task with/without replay.  
(3)~\textbf{IncLoRA} allows incremental learning of new LoRA
parameters for each task without constraints.  
(4)~\textbf{SeqSVD} uses fixed-rank SVD adapters trained sequentially without RorR.   
(5)~\textbf{EWC}~\cite{kirkpatrick2017overcoming} finetunes LoRA with
a regularization loss to prevent interference with past tasks.  
(6)~\textbf{LwF}~\cite{li2017learning} distills the model of
the last task using the current task data.  
(7)~\textbf{L2P}~\cite{wang2022learning} instantiates a prompt pool for adaptive prompt selection and prompt tuning for individual samples.  
(8)~\textbf{LFPT5}~\cite{qin2021lfpt5} learns soft prompts and generates pseudo-data for replay.  
(9)~\textbf{L-CL} trains SVD adapters incrementally with SVD regularization.  
(10)~\textbf{B-CL} applies SVD regularization with gradient projection.  
(11)~\textbf{O-LoRA}~\cite{wang2023orthogonal} enforces orthogonality between LoRA updates across tasks. 
(12) ~\textbf{MIGU}~\cite{du2024unlocking} updates parameters based on gradient magnitude.
(13)~\textbf{LB-CL}~\cite{qiao2024learn} initializes low-rank matrix parameters in
new tasks from specific past parameters besides enforcing orthogonality.  
(14)~\textbf{DATA}~\cite{liao2025data} decomposes attention into high and low rank subspaces, and leverages orthogonality with optional replay.  
(15)~\textbf{Recurrent KIF}~\cite{feng2025recurrent} uses an expensive recurrent mechanism to modulate task-specific adapter reuse using replay.
\begin{table*}[htbp]
\centering
\begin{tabular}{ll}
\toprule
\textbf{Task} & \textbf{Prompts} \\
\midrule
NLI & What is the logical relationship between the "sentence 1" and the "sentence 2"? \\
    & Choose one from the option. \\
QQP & Whether the "first sentence" and the "second sentence" have the same meaning? \\
    & Choose one from the option. \\
SC  & What is the sentiment of the following paragraph? Choose one from the option. \\
TC  & What is the topic of the following paragraph? Choose one from the option. \\
BoolQA & According to the following passage, is the question true or false? \\
       & Choose one from the option. \\
MultiRC & According to the following passage and question, is the candidate answer true \\
        & or false? Choose one from the option. \\
WiC & Given a word and two sentences, whether the word is used with the same sense \\
    & in both sentence? Choose one from the option. \\
FOMC & What is the monetary policy stance for the following text? \\
& Choose one from the option. \\
20Minuten & Provide a simplified version of the following paragraph in German.\\
ScienceQA & Choose an answer for the following question and give your reasons. \\
NumGLUE-cm & Solve the following math problem. \\
NumGLUE-ds & Solve the following math problem. \\
Py150 & Continue writing the code.\\
MeetingBank & Write a summary of the following meeting transcripts.\\
C-STANCE & Determine the attitude of the following text towards the specified object. \\ 
&Choose one from the option. \\

\bottomrule
\end{tabular}
\caption{Instructions for different tasks.}
\label{tab:task_prompts}
\end{table*}
\subsection{Instruction Tuning.} Instruction-following is a fundamental capability for LLMs to serve as an effective interface between humans and AI systems~\cite{wei2021finetuned,ouyang2022training}. We adopt instruction tuning as our training paradigm for two key reasons: (1) it enables the incorporation of human prior knowledge via explicit instructions, thereby facilitating more efficient learning; and (2) it enhances generalization by guiding the model to learn underlying principles that apply across tasks.

All tasks are formatted using a unified schema consisting as follows: (1) \textit{Task Instruction} -- a clear description of how to map an input (e.g., a sentence or document) to an output; (2) \textit{Options} -- the constrained set of valid output labels for the task; (3) \textit{Text} -- the input instance provided to the model; and (4) \textit{Answer} -- the correct target output. This combined sequence is fed into the pretrained model to guide prediction for all experiments.
\subsection{Task Instructions}
Table~\ref{tab:task_prompts} shows the prompts used for different tasks. NLI
denotes natural language inference, and includes tasks
MNLI, RTE and CB. SC denotes sentiment
analysis, including Amazon, Yelp, SST-2 and
IMDB. TC denotes topic classification and contains the tasks
AG News, Dbpedia and Yahoo.

\subsection{Implementation Details} 
\label{app:implementation_details}
Our implementation for ELLA uses PyTorch v2.0.1~\cite{paszke2019pytorch} and Deepspeed v0.10.0~\cite{rasley2020deepspeed}.
Our experiments were conducted on machines equipped with $4$ A$40$ GPUs/ $8$ V$100$ GPUs for ELLA and all baselines, except DATA/DATA+Replay, which required $2$ H$100$ GPUs. For all orders of task
streams for the Standard-CL Benchmark and the Long Sequence Benchmark, we trained the models with one epoch using the AdamW optimizer~\cite{loshchilov2017decoupled} and WarmupLR scheduler~\cite{deepspeed_schedulers} with a total batch size of $32$. We used a constant learning rate of $1e-3$ for T$5$ and $1e-4$ for LLaMA, a dropout rate of
$0.1$, and a weight decay rate of $0$. For TRACE Order $7$ (C-STANCE, FOMC,
MeetingBank, Py150, ScienceQA, NumGLUE-cm,
NumGLUE-ds, 20Minuten), we trained with $5000$
samples $5$, $3$, $7$, $5$, $3$, $5$, $5$, $7$ epochs respectively. The coefficient $\lambda$ for different task orders has been reported in Table~\ref{tab:lambda_settings}. $\lambda$ was selected based on performance on a small, held-out validation set. To establish an effective search range, we followed the common heuristic of choosing values that scale the regularization term, $\mathcal{L}_{\text{ELLA}}$ (promoting stability), to a similar order of magnitude as the current task accuracy loss (promoting plasticity) for balanced learning. We observed that performance was robust across this range and that a single $\lambda$ often generalized well across multiple subsequent tasks. The LoRA
rank was set to $8$ for all experiments, and the proportion of past task data mixed in for replay methods was set to $2\%$ of the original training set. 


\begin{table*}[ht]
\centering
\resizebox{0.9\textwidth}{!}{%
\begin{tabular}{cll}
\toprule
\textbf{Order} & \textbf{T$5$ $\lambda$} & \textbf{LLaMA $\lambda$} \\
\midrule
1--3 & $0, 3\times10^{4},\,\ldots,\, 3\times10^{4}$ & $0, 3\times10^{6},\,\ldots,\,3\times10^{6}$ \\
\midrule
4 & $0,\,5{\times}10^{5},\,\ldots,\,5{\times}10^{5},\,5{\times}10^{7}$ 
  & $0,\,5{\times}10^{8},\,\ldots,\,5{\times}10^{8}$ \\
\midrule
5 & $0,\,5{\times}10^{6},\,\ldots,\,5{\times}10^{6},\,5{\times}10^{7},\,5{\times}10^{7},\,5{\times}10^{7}$ 
  & $0,\,5{\times}10^{6},\,\ldots,\,5{\times}10^{6},\,5{\times}10^{7},\,5{\times}10^{7},\,5{\times}10^{7}$ \\
\midrule
6 & $0,\,5{\times}10^{5},\,\ldots,\,5{\times}10^{5}$ 
  & $0,\,5{\times}10^{8},\,\ldots,\,5{\times}10^{8}$ \\
\midrule
7 & $0,\,5{\times}10^{5},\,\ldots,\,5{\times}10^{5}$ 
  & $0,\,5{\times}10^{7},\,5{\times}10^{7},\,5{\times}10^{7},\,5{\times}10^{9},\,5{\times}10^{9},\,5{\times}10^{9},\,5{\times}10^{9}$ \\
\bottomrule
\end{tabular}}
\caption{Coefficient $\lambda$ settings for different task orders in T$5$ and LLaMA.}
\label{tab:lambda_settings}
\end{table*}

\section{Artifact Licenses}
\label{license}
According to their license, all the LLMs used in this paper fall under acceptable use cases. The licenses are listed for perusal: T$5$-Base (\url{https://huggingface.co/google-t5/t5-base}), T$5$-Large (\url{https://huggingface.co/google-t5/t5-large}), T$5$-XL (\url{https://huggingface.co/google/t5-v1_1-xl}), LLaMA-3.1-8b (\url{https://huggingface.co/meta-llama/Llama-3.1-8B/blob/main/LICENSE}).


\end{document}